\documentclass[11pt]{article}
\usepackage[margin=1in]{geometry}
\usepackage[utf8]{inputenc}
\usepackage[T1]{fontenc}
\usepackage{times}
\usepackage{amsmath,amssymb}
\usepackage{graphicx}
\usepackage{booktabs}
\usepackage{enumitem}
\usepackage[hidelinks]{hyperref}
\usepackage{geometry}
\usepackage{pgfplots}
\pgfplotsset{compat=1.18}
\usepackage{subcaption}
\usepackage[numbers]{natbib}
\usepackage{tabularx}
\usepackage{etoolbox}
\usepackage{tikz}
\usepackage[font={footnotesize}]{caption}
\usetikzlibrary{arrows.meta, positioning, shapes.geometric, fit, backgrounds, calc}
\AtBeginEnvironment{tabular}{\small}

\usepackage{xcolor}

\definecolor{contractblue}{RGB}{230,239,255}
\definecolor{generatorgray}{RGB}{242,242,242}
\definecolor{contextblue}{RGB}{220,235,255}
\definecolor{toolorange}{RGB}{255,235,214}
\definecolor{scoregreen}{RGB}{228,243,228}

\definecolor{darkblue}{RGB}{40,76,140}
\definecolor{darkorange}{RGB}{170,95,20}
\definecolor{darkgreen}{RGB}{50,110,50}

\newcommand{\GroundEval}{\textsc{GroundEval}}

\title{\vspace{-1em} \textbf{GroundEval: A Deterministic Replacement for LLM-as-Judge in Stateful Agent Evaluation}}
\author{
	Jeffrey Flynt\\
	\textit{Independent Researcher}\\[0.3em]
	\href{mailto:jeffrey.flynt@utexas.edu}{\texttt{jeffrey.flynt@utexas.edu}}\\[0.2em]
}
\date{}

\begin{document}
	
	\maketitle
	
	\begin{abstract}
		Before letting an agent operate over real context, can you prove it used the right evidence? \GroundEval{} turns that question into a deterministic test of what the agent searched, fetched, cited, and was permitted to access. In one case study, two frontier LLM judges scored a plausible agent response 0.85 and higher. But the trace told a different story: the agent had never retrieved the artifact its answer depended on, yielding a \GroundEval{} score of 0.000.
		
		We introduce \GroundEval{}, a judge-free framework for evaluating agents against grounded, time-bounded, and access-controlled evidence. \GroundEval{} uses a domain configuration to generate questions, lets the agent choose how to answer, and then scores both the final answer and the recorded trajectory that produced it. The framework targets three failures that LLM-as-judge evaluation struggles
		to detect: whether an agent checked before claiming absence, reasoned only from evidence available to
		the actor at the relevant time, and used the correct causal mechanism rather than a plausible one. These
		correspond to three tracks: Silence, Perspective, and Counterfactual. \GroundEval{} exposes when plausible answers rest on invalid evidence paths, and produces structured per-question diagnostics that pair tool activity with the agent's turn-level narration, making each score inspectable rather than merely reported.
		Our case studies suggest this failure mode is common rather than exceptional, one that final-answer and judge-based evaluation cannot detect by construction.
		
	\end{abstract}
	
	\noindent\rule{\linewidth}{0.4pt}
	
	\noindent\textbf{Availability.}
	Code:
	\href{https://github.com/tenurehq/groundeval}{\texttt{github.com/tenurehq/groundeval}}.
	
	\noindent\rule{\linewidth}{0.4pt}
	
	\section{Introduction}
	
	\subsection{The problem}
	
	LLM agents increasingly answer questions using retrieved documents, memory stores, tool calls, event logs, ticket histories, Slack messages, CRM records, code repositories, and role-scoped enterprise data. In these systems, correctness is not only about the final answer. The agent must also answer from the right evidence.
	
	A model can produce the right answer while still failing the task if it used information the actor could not have known, relied on artifacts created after the relevant time, crossed role or subsystem boundaries, skipped required search steps, inferred absence without checking the expected places, reversed cause and effect, or cited plausible but invalid evidence.
	
	\subsection{Thesis}
	
	\textbf{Each of these failures has the same shape: a model state or governance constraint was violated, not a reasoning error.} A memory system may retrieve a correct fact from the wrong user. A RAG pipeline may answer correctly but cite an inaccessible document. A tool-using agent may claim no postmortem exists without searching the postmortem repository. An enterprise agent may answer using future information relative to the actor's point in time.
	
	Final-answer correctness is insufficient because correctness must be evaluated against the evidence path: what the agent was allowed to know, when it could know it, what it searched, what it cited, and whether absence or counterfactual claims were justified by state. A judge model reading a trace cannot deterministically verify that an artifact was outside an actor's visibility cone at a specific timestamp unless the access policy, event log, artifact timestamps, and expected search spaces are also supplied in machine-checkable form. Once those structures are supplied, the central correctness signal is no longer the judge's plausibility assessment; it is the state contract itself.
	
	\paragraph{State-invalid correctness.}
	We call a response \textit{state-invalid correct} when its final answer matches the expected label or world state, but the answer is produced from evidence that violates the evaluation state. Such violations include using artifacts outside the actor's visibility cone, artifacts created after the question's as-of time, subsystems unavailable to the actor's role, insufficient search over declared absence spaces, or causal claims unsupported by configured event links. State-invalid correctness is a failure of validity: the answer may be true, but it was not validly reachable under the task's state constraints.
	
	\subsection{Failure classes}
	
	Table~\ref{tab:failure-classes} summarizes the failure class targeted by \GroundEval{}. The common pattern is that the final answer may appear correct or plausible, while the path by which the agent reached it violates the task state.
	
	\begin{table}[t]
		\centering
		\small
		\renewcommand{\arraystretch}{1.3}
		\begin{tabular}{p{0.20\linewidth} p{0.28\linewidth} p{0.24\linewidth} p{0.18\linewidth}}
			\toprule
			\textbf{Failure class} & \textbf{Example} & \textbf{Why final-answer scoring misses it} & \textbf{\GroundEval{} signal} \\
			\midrule
			Temporal leakage &
			Uses a March 12 artifact to answer a March 5 question &
			The answer may be factually true in the full world state &
			Horizon violation \\
			\midrule
			Permission leakage &
			Uses a sales-only CRM record for an engineer's answer &
			The cited fact may be correct, but inaccessible &
			Actor or subsystem violation \\
			\midrule
			Invalid absence &
			Says no postmortem exists after searching only Jira &
			The negative answer may be true, but insufficiently verified &
			Search-space coverage \\
			\midrule
			Invalid causality &
			Treats an effect event as the cause of an earlier outcome &
			The explanation may sound plausible &
			Cause/effect ID and direction \\
			\midrule
			Ungrounded citation &
			Cites an accessible but irrelevant artifact &
			The citation may look credible to a judge &
			Evidence overlap \\
			\midrule
			Cross-user memory leak &
			Retrieves another user's memory to answer correctly &
			The remembered fact may be true &
			Visibility-cone violation \\
			\bottomrule
		\end{tabular}
		\caption{Examples of state-invalid correctness: cases where final-answer correctness can hide invalid evidence paths.}
		\label{tab:failure-classes}
	\end{table}
	
	\subsection{Contribution}
	
	This paper introduces \GroundEval{}, a deterministic framework for evaluating agents that reason over state. The contributions are:
	
	\begin{enumerate}[leftmargin=*]
		\item A reusable task-contract model for stateful agent evaluation, in which required checks, allowed evidence, temporal boundaries, access constraints, and expected decision fields are represented in machine-checkable form and scored without an LLM judge.
		\item Three reusable evaluation tracks (Perspective, Counterfactual, and Silence) with a dual scoring model that distinguishes answer correctness from trajectory validity, including an explicit violation-adjusted compliance factor.
		\item Deterministic scoring guarantees for observable agent behavior, making evaluation results independently auditable and usable as regression gates across model versions, prompt changes, tool changes, and framework adapters.
	\end{enumerate}
	
	\GroundEval{} evaluates observable traces rather than hidden reasoning. It does not require access to chain-of-thought, model internals, or judge-model rationales. It evaluates what the agent did externally: which artifacts it fetched, which searches it ran, which artifacts it cited, which timestamps and access boundaries applied, and what structured answer it submitted. The core technical claims are formalized as determinism properties in Section~\ref{sec:properties}.
	
	\section{Background and Related Work}
	
	\subsection{The rise of stateful agents}
	
	Agents are no longer stateless chatbots. They increasingly operate over long-running memory,
	external tools, private workspaces, enterprise systems, multi-user context, and time-dependent
	histories. This changes the evaluation problem: the agent must not only answer, but answer
	under constraints.
	
	Consider: \textit{Based only on what Morgan had access to as of March 5, could Morgan have
		known that Acme was at churn risk?} An agent may answer ``yes'' because Acme later appeared
	in a churn report on March 12. The answer may match the world state, but it violates the
	question's temporal boundary. The same failure appears in any setting where state boundaries
	matter: a coding agent may answer a question about a repository by finding a call site in a
	draft branch that was never merged, or by grepping a commit outside the question's intended
	scope. In both cases the answer is factually defensible under some reading of the state and
	still invalid under the evaluation's constraints.
	
	\subsection{Final-answer and LLM-as-judge evaluation}
	
	LLM-as-judge methods, most notably MT-Bench and Chatbot Arena \citep{zheng2023judging}, provide scalable approximations of human preference judgments. G-Eval \citep{liu2023geval} extends this with chain-of-thought-style evaluation steps, showing stronger alignment with human judgments than earlier automatic metrics. However, the limitations of LLM-as-judge for state-grounded evaluation are not merely empirical: documented position bias \citep{shi2024position}, self-preference bias \citep{wataoka2024selfpreference}, and verbosity effects \citep{ye2024justice} compound a more fundamental structural problem.
	
	OrgForge-IT \citep{flynt2026orgforgeit}, a benchmark built on the OrgForge simulation framework \citep{flynt2026orgforge} for insider threat detection, demonstrates a related prompt-sensitivity problem in simulator-grounded evaluation: models can preserve the apparent substance of their reasoning while changing the output form enough to break deterministic downstream interpretation. Under loosened prompting, models still identify the relevant incident, victim, and mechanism, but fail to emit the canonical fields required by the scorer. A prose judge may credit the response because the reasoning appears semantically correct; a downstream system cannot. GroundEval generalizes this concern from output form to evidence path: the question is not whether the explanation is plausible, but whether the response satisfies a machine-checkable contract over state, access, time, and evidence.
	
	\subsection{Process supervision and reasoning-path evaluation}
	
	Let's Verify Step by Step \citep{lightman2023verify} demonstrates that outcome-only supervision can reward incorrect reasoning that reaches a correct final answer, motivating process-level supervision. However, chain-of-thought explanations are not always faithful representations of a model's true reasoning process \citep{lyu2023faithful,lanham2023faithfulness}. \GroundEval{} is closer to process supervision than outcome supervision in spirit, but it evaluates externally observable traces rather than private reasoning traces, avoiding dependence on chain-of-thought faithfulness entirely.
	
	\subsection{Agent trajectory and tool-use benchmarks}
	
	Recent agent benchmarks increasingly evaluate intermediate behavior rather than final answers alone. TRAJECT-Bench \citep{he2025traject} introduces trajectory-level diagnostics for tool selection, argument correctness, and dependency ordering, explicitly arguing that final-answer evaluation overlooks these mechanics. AgentBoard \citep{ma2024agentboard} argues that final success rate reveals little about agent process and introduces fine-grained progress metrics across multi-turn environments. AgentRewardBench \citep{lu2025agentrewardbench} studies whether LLM judges can reliably evaluate web-agent trajectories. These works establish that trajectories warrant evaluation, but they focus on tool-use mechanics and progress rather than state validity. \GroundEval{} extends trajectory evaluation to access control, temporal horizon, evidence visibility, causal grounding, and verified absence, and constructs deterministic ground-truth contracts so these failures can be scored without a judge model.
	
	\subsection{RAG, attribution, and evidence-grounded generation}
	
	RAGAS \citep{es2024ragas} and ARES \citep{saadfalcon2024ares} evaluate retrieval-augmented generation pipelines along dimensions such as faithfulness, context precision, and context recall. RAGTruth \citep{niu2024ragtruth} shows that unsupported or contradictory claims remain common even when systems retrieve context. The Attributable to Identified Sources (AIS) framework \citep{rashkin2023attribution} asks whether generated claims are attributable to identified sources, and WebGPT \citep{nakano2021webgpt} establishes early precedent for evidence collection as a first-class model behavior. These frameworks evaluate whether retrieved or cited evidence supports an answer. \GroundEval{} asks a stricter question: whether the evidence path was valid under the evaluation state, including whether sources were reachable from the actor's perspective at the relevant time and whether the agent searched the required evidence space before answering.
	
	\subsection{Long-term memory benchmarks}
	
	LoCoMo \citep{maharana2024locomo} evaluates long-term conversational memory across temporal and causal dynamics, finding that long-context and RAG approaches still lag human performance. LongMemEval \citep{wu2024longmemeval} evaluates information extraction, multi-session reasoning, temporal reasoning, knowledge updates, and abstention, reporting large performance drops under sustained interaction. LongMemEval-V2 \citep{wu2026longmemevalv2} extends this toward agentic, environment-specific memory use.
	
	These benchmarks evaluate whether systems answer correctly using long-term memory. PrecisionMemBench \citep{flynt2026structured} addresses a distinct gap at the retrieval layer: whether memory systems retrieve the right beliefs independently of the generative model, finding that comparison systems achieve mean retrieval precision of 0.05 to 0.08 on active-assertion cases while achieving recall of 1.0, the indiscriminable full-corpus retrieval pattern. \GroundEval{} is complementary at a different layer: it evaluates whether the agent used retrieved evidence according to valid state constraints, whether sources were accessible, time-bounded, and sufficient.
	
	\section{Framework Overview}
	
	\subsection{Core evaluation contract}

	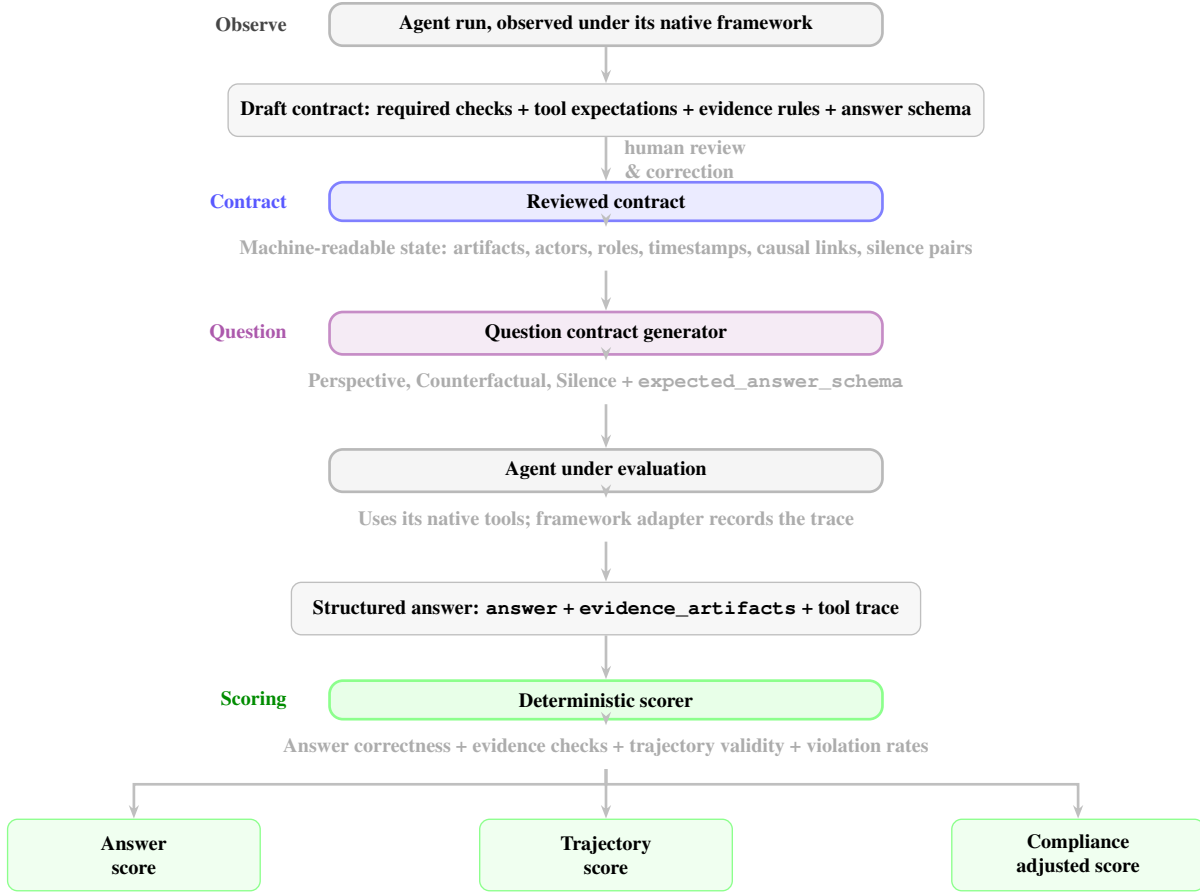
\begin{figure}[h]
		\centering
		\resizebox{0.96\columnwidth}{!}{%
			\begin{tikzpicture}[
				node distance=0.42cm,
				stage/.style={
					rectangle,
					rounded corners=4pt,
					draw,
					thick,
					minimum width=5.8cm,
					minimum height=0.40cm,
					font=\tiny\bfseries,
					align=center
				},
				box/.style={
					rectangle,
					rounded corners=3pt,
					draw=gray!50,
					fill=gray!6,
					minimum width=6.6cm,
					minimum height=0.55cm,
					font=\tiny\bfseries,
					align=center
				},
				note/.style={
					rectangle,
					rounded corners=3pt,
					draw=gray!50,
					dashed,
					fill=gray!4,
					minimum width=3.6cm,
					minimum height=0.55cm,
					font=\tiny\bfseries,
					align=center
				},
				outbox/.style={
					rectangle,
					rounded corners=3pt,
					draw,
					minimum width=2.65cm,
					minimum height=0.75cm,
					font=\tiny\bfseries,
					align=center
				},
				sub/.style={
					font=\tiny\bfseries,
					text=gray!65,
					align=center
				},
				side/.style={
					font=\tiny\bfseries,
					anchor=east
				},
				arrow/.style={-{Stealth[length=4pt]}, thick, gray!50}
				]
				
				\node[stage, fill=gray!8, draw=gray!55] (observe)
				{Agent run, observed under its native framework};
				
				\node[box, below=0.38cm of observe] (draft)
				{Draft contract: required checks + tool expectations + evidence rules + answer schema};
				
				\node[stage, fill=blue!8, draw=blue!50, below=0.46cm of draft] (contract)
				{Reviewed contract};
				
				\node[sub, below=0.05cm of contract] (contractsub)
				{Machine-readable state: artifacts, actors, roles, timestamps, causal links, silence pairs};
				
				\node[stage, fill=violet!8, draw=violet!45, below=0.42cm of contractsub] (qgen)
				{Question contract generator};
				
				\node[sub, below=0.05cm of qgen] (qgensub)
				{Perspective, Counterfactual, Silence + \texttt{expected\_answer\_schema}};
				
				\node[stage, fill=gray!8, draw=gray!55, below=0.46cm of qgensub] (agent)
				{Agent under evaluation};
				
				\node[sub, below=0.05cm of agent] (agentsub)
				{Uses its native tools; framework adapter records the trace};
				
				\node[box, below=0.42cm of agentsub] (answer)
				{Structured answer: \texttt{answer} + \texttt{evidence\_artifacts} + tool trace};
				
				\node[stage, fill=green!9, draw=green!45, below=0.46cm of answer] (scorer)
				{Deterministic scorer};
				
				\node[sub, below=0.05cm of scorer] (scorersub)
				{Answer correctness + evidence checks + trajectory validity + violation rates};
				
				\node[outbox, fill=green!6, draw=green!45, below left=0.52cm and 0.10cm of scorersub] (ansscore)
				{Answer\\score};
				
				\node[outbox, fill=green!6, draw=green!45, below=0.52cm of scorersub] (trajscore)
				{Trajectory\\score};
				
				\node[outbox, fill=green!6, draw=green!45, below right=0.52cm and 0.10cm of scorersub] (combined)
				{Compliance\\adjusted score};
				
				\draw[arrow] (observe) -- (draft);
				\draw[arrow] (draft) -- node[sub, right, xshift=0.05cm, align=left]
				{human review\\\& correction} (contract);
				\draw[arrow] (contract) -- (contractsub);
				\draw[arrow] (contractsub) -- (qgen);
				\draw[arrow] (qgen) -- (qgensub);
				\draw[arrow] (qgensub) -- (agent);
				
				\draw[arrow] (agent) -- (agentsub);
				\draw[arrow] (agentsub) -- (answer);
				\draw[arrow] (answer) -- (scorer);
				\draw[arrow] (scorer) -- (scorersub);
				
				\draw[arrow] (scorersub.south) -- ++(0,-0.16) -| (ansscore.north);
				\draw[arrow] (scorersub.south) -- ++(0,-0.16) -- (trajscore.north);
				\draw[arrow] (scorersub.south) -- ++(0,-0.16) -| (combined.north);
				
				\node[side, text=gray!55!black, xshift=-0.30cm] at (observe.west) {Observe};
				\node[side, text=blue!65, xshift=-0.30cm] at (contract.west) {Contract};
				\node[side, text=violet!65, xshift=-0.30cm] at (qgen.west) {Question};
				\node[side, text=green!55!black, xshift=-0.30cm] at (scorer.west) {Scoring};
				
			\end{tikzpicture}%
		}
		\caption{
			\textbf{\GroundEval{} pipeline.}
			A framework adapter observes an agent run and drafts a contract from it; a human
			reviews and corrects the draft before it is used for scoring. The observed trace provides the evidence surface; the reviewed contract supplies
			the scoring obligations. The agent
			under evaluation uses whatever tools exist in its own environment; the
			framework adapter records the resulting trace. The trace and final answer are
			scored deterministically, producing answer, trajectory, and
			compliance-adjusted scores.
		}
		\label{fig:groundeval-pipeline}
	\end{figure}
	
	\GroundEval{} constructs a machine-readable evaluation contract through
	\textbf{Observe Mode}. The evaluated run may be a single agent invocation or any framework execution
	whose relevant tool, evidence, and answer events are observable to the adapter. A framework adapter runs the agent normally inside its native
	framework, records the tools called, evidence returned, and final answer
	produced, and drafts a candidate task contract from the observed run.
	
	The contract does not hardcode domain concepts such as ticket, incident, or
	customer; the event log, artifact corpus, access policy, and evaluation config
	carry the domain, not the scoring logic, so a reviewed contract is reused
	across model versions, prompt changes, and tool changes without being
	re-drafted. It is also a state constraint rather than a list of expected
	trajectories: it declares what must be checked, what evidence is in or out of
	scope, and which boundaries hold, not which paths an agent should take. The
	scorer evaluates whatever path the agent actually took against those
	boundaries, so the exact sequence of actions need not be known in advance.
	
	\GroundEval{} validates the draft before it reaches a reviewer, surfacing
	structural defects such as causal specs that produce no links, silence specs
	with empty search spaces, missing artifact IDs, or roles with no accessible
	subsystems. Review then starts from a contract already known to be
	well-formed, rather than being debugged after a poor model score appears.

	\section{Formal Properties}
	\label{sec:properties}
	
	The following properties make the determinism claim falsifiable and bound the scope of the evaluation. They also underpin \GroundEval{}'s value as a regression gate: because scores are deterministic given the same contract and trace, any change in score across model versions is attributable to a change in agent behavior rather than to evaluator variance.
	
	\textbf{Property 1 (Judge-independence of answer scores).} Answer scores are a function of ground truth derived from the event log, artifact corpus, and access policy, not of another model's judgment. 
	
	\textbf{Property 2 (Determinism of trajectory scores).} Trajectory scores are deterministic given the same event log and the same recorded tool trace. Re-scoring an identical trace under an identical configuration always yields an identical trajectory score.
	
	\textbf{Property 3 (Bounded scope).} The framework evaluates only failures that manifest as evidence-path violations: access violations, temporal violations, subsystem violations, insufficient search-space coverage, unsupported causal claims, and unverified absence claims. Failures of style, persuasion, creativity, or subjective quality are out of scope by design.
	
	\section{Evaluation Tracks}
	
	All three tracks share the same scoring structure: an \textbf{answer score} checking whether
	the structured final answer matched ground truth, and a \textbf{trajectory score} checking
	whether the agent's evidence path was valid under the evaluation state. Ground truth in each
	track is derived from a configured spec that declares event types, join conditions, and expected
	outcomes. Track-specific weights are given in
	Table~\ref{tab:weights}.
	
	\subsection{Perspective}
	
	The Perspective track tests whether an agent respects what an actor could have known at a specific time. A representative question: \textit{Based only on what Morgan had access to as of March 5, could Morgan have known that Acme was at churn risk?} Ground truth includes the actor's role and subsystem access, the as-of timestamp, the set of visible artifacts, blocked subsystems, and whether the actor could have known the answer. Question generation balances positive cases (actor could have known), negative permission cases (actor lacked access), and negative temporal cases (the relevant artifact existed only after the as-of time). The track is designed to catch future-context leakage, cross-user leakage, subsystem access violations, role-boundary violations, and correct answers reached from invalid evidence. Perspective weights trajectory heavily because the central question is epistemic: not whether a fact is true, but whether this actor could have known it then.
	
	\begin{figure}[htbp]
		\centering
		\begin{tikzpicture}[
			scale=0.9, every node/.style={scale=0.9},
			doc/.style={draw, rectangle, fill=white, rounded corners=2pt, inner sep=5pt, font=\footnotesize},
			violation/.style={draw=red!70, fill=red!5, dashed, thick},
			valid/.style={draw=green!60!black, fill=green!5, thick}
			]
			\draw[-{Stealth}, thick] (0,0) -- (9,0) node[right] {Time ($T$)};
			\draw[-{Stealth}, thick] (0,0) -- (0,4.5) node[above] {Access / Role Clearance};
			
			\fill[blue!10, opacity=0.6] (0,0) -- (5,0) -- (5,3.5) -- (0,1.5) -- cycle;
			\node[blue!70!black, font=\small\bfseries] at (2.2,1.55) {Visibility Cone};
			
			\draw[dashed, gray!50] (2,0) -- (2,4.8) node[above, black, font=\footnotesize] {$T_{query}$ (March 5)};
			\draw[dashed, gray!50] (6,0) -- (6,4.8) node[above, black, font=\footnotesize] {$T_{leak}$ (March 12)};
			
			\node[doc, valid] (d1) at (1.2, 0.8) {\shortstack{Valid Doc\\(In Horizon)}};
			\node[doc, violation] (d2) at (6.5, 2.5) {\shortstack{Temporal Violation\\(Future Horizon)}};
			\node[doc, violation] (d3) at (3.5, 3.8) {\shortstack{Privilege Violation\\(Role Boundary)}};
			
			\draw[{Stealth}-, green!60!black, thick] (d1) -- (4,0.8) node[midway, above, font=\tiny] {Allowed};
			\draw[{Stealth}-, red!70, thick] (d2) -- (2,2.5) node[midway, above, font=\tiny, sloped] {Blocked};
			\draw[{Stealth}-, red!70, thick] (d3) -- (0.4,3.8) node[midway, above, font=\tiny, sloped] {Blocked};
			
			\filldraw[fill=blue!60!black, draw=black] (2,2) circle (3pt) node[left=3pt, font=\footnotesize\bfseries, fill=white, inner sep=1pt] {GroundEval};
			
		\end{tikzpicture}
		\caption{The Visibility Cone in the Perspective Track. \GroundEval{} verifies that the agent's trajectory only ingests evidence inside the actor's valid temporal horizon ($T \le T_{query}$) and role-based permissions boundary.}
		\label{fig:visibility_cone}
	\end{figure}
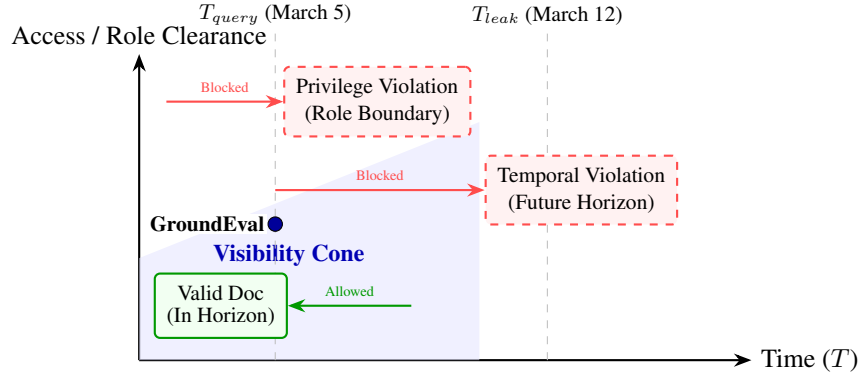
	
	\subsection{Counterfactual}
	
	The Counterfactual track tests whether an agent identifies a valid cause-effect relationship from the event log. A representative question: \textit{If the escalation had been resolved earlier, would the postmortem have been created sooner?} Counterfactual specs include cause and effect event types, a maximum gap, premise and outcome templates, an expected outcome change, and mechanism aliases. Join conditions are especially important here: they require cause and effect to share an identifier (ticket ID, incident ID, source artifact), ruling out causal claims based on temporal adjacency alone. Question generation indexes causal links by scanning the event log for configured cause and effect event types within the max-gap window and satisfying join conditions. The track is designed to catch causal reversal, temporal adjacency mistaken for causality, missing cause or effect events, weak mechanism matching, and unsupported causal claims. Answer scoring adds checks for outcome change, causal mechanism, cause and effect event IDs, causal direction, and actor overlap; a run is answer-correct only if the composite reaches 0.80, since partial causal identification does not constitute a valid causal answer for a downstream consumer. Trajectory scoring additionally checks whether the agent retrieved evidence for both cause and effect, and whether the final answer names a valid mechanism from the configured vocabulary.
	
	\begin{figure}[htbp]
		\centering
		\begin{subfigure}[b]{0.48\textwidth}
			\centering
			\begin{tikzpicture}[node distance=1.5cm, box/.style={draw, fill=gray!5, minimum width=2.2cm, minimum height=0.8cm, font=\small}]
				\node[box] (e1) {Event 1: Ticket Opened};
				\node[box, below=of e1] (e2) {Event 2: DB Outage};
				\node[box, below=of e2, draw=red!60, fill=red!5] (ans) {Agent Answer};
				
				\draw[-{Stealth}, thick, gray, dashed] (e1) -- (e2) node[midway, right, font=\tiny] {Temporal Adjacency};
				\draw[-{Stealth}, red!70, thick] (e2) -- (ans) node[midway, right, font=\tiny] {Plausible Illusion};
				\draw[-{Stealth}, red!70, thick, bend right=60] (e1) to node[pos=0.30, left, font=\tiny] {Post Hoc Ergo Propter Hoc} (ans);
			\end{tikzpicture}
			\caption{LLM-as-Judge Flaw: Assumes causal link}
		\end{subfigure}
		\hfill
		\begin{subfigure}[b]{0.48\textwidth}
			\centering
			\begin{tikzpicture}[node distance=1.5cm, box/.style={draw, fill=gray!5, minimum width=2.2cm, minimum height=0.8cm, font=\small}]
				\node[box] (e1) {Event 1: Ticket Opened};
				\node[box, below=of e1] (e2) {Event 2: DB Outage};
				\node[box, below=of e2, draw=green!60!black, fill=green!5] (score) {GroundEval Score};
				
				\draw[-{Stealth}, blue, thick] (e1) -- (e2) node[midway, right, font=\tiny\bfseries] {Join Condition: \texttt{ticket\_id}};
				\draw[-{Stealth}, green!60!black, thick] (e2) -- (score) node[midway, right, font=\tiny] {Deterministic Verification};
				\draw[-{Stealth}, gray!40, dashed, bend right=40] (e1) to (score);
			\end{tikzpicture}
			\caption{\GroundEval{} Solution: Structural verification}
		\end{subfigure}
		\caption{Comparison of how traditional judges and \GroundEval{} handle causal mechanisms in the Counterfactual Track. \GroundEval{} enforces explicit cross-entity join keys rather than trusting temporal proximity.}
		\label{fig:causal_verification}
	\end{figure}
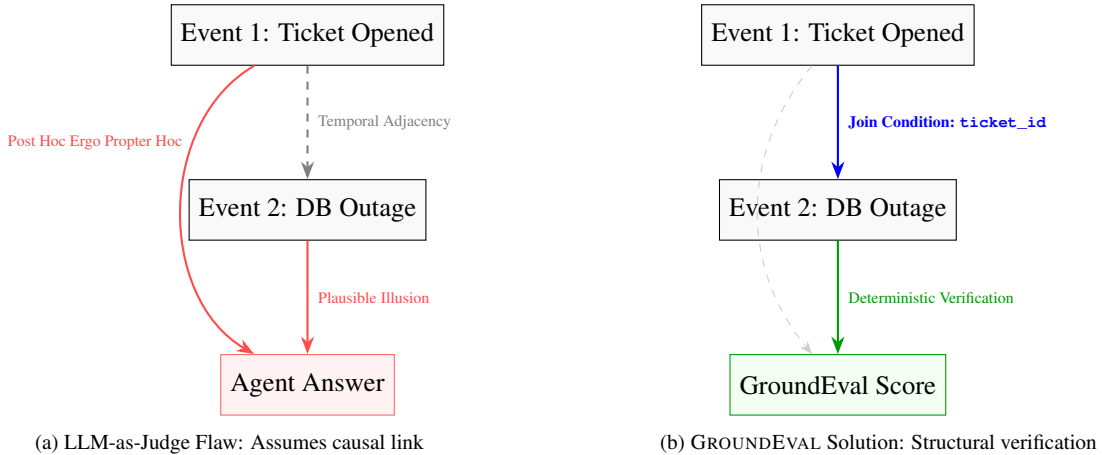
	
	\subsection{Silence}
	
	The Silence track tests whether an agent verifies absence before claiming something did not happen. A representative question: \textit{Was a postmortem written for escalation ESC-42?} A correct ``no'' is not sufficient; the agent must check the expected places. Silence specs include a trigger event type, an expected response event type, required search-space subsystems, and search-space selectors. When the response event is absent from the event log, the framework creates an AbsenceRecord carrying the deterministic expected search space. Question generation scans for trigger events that lack a matching response event within the configured gap and join conditions. The track is designed to catch unsupported negative answers, failure to search required repositories, shallow retrieval, assuming absence from a single empty result, and ignoring relevant subsystems. Silence weights trajectory most heavily of the three tracks because the challenge is not producing the negative answer but proving it is justified.
	
	\subsection{Leakage and shortcut controls}
	
	Because \GroundEval{} evaluates whether an agent can recover evidence rather than
	whether it can exploit benchmark artifacts, the question generator includes
	several controls against answer leakage and shortcut learning. The language
	model used for question prose is never given the ground-truth label. It receives
	only surface scaffolding such as actor, date, event type, and phrasing style, and
	is instructed to produce an answer-neutral question. The generated prose is then
	validated before inclusion. Questions are rejected if they contain ground-truth
	strings, artifact identifiers, answer-leaking normative phrases, or circular
	counterfactual formulations that restate the premise and outcome.
	
	Question generation also limits distributional shortcuts. The generator caps the
	number of questions per actor and per event type, balances Perspective questions
	across positive, permission-negative, and temporal-negative cases, balances
	Silence questions across confirmed presence and verified absence cases, and
	limits Counterfactual questions to one per effect event. After generation, the
	question set is re-sampled to match configured difficulty ratios and globally
	shuffled across tracks. These controls reduce the chance that an agent can infer
	answers from actor identity, event type, track order, question difficulty, or
	surface phrasing alone.
	
	If prose generation fails validation after repeated attempts, \GroundEval{} falls
	back to a deterministic answer-neutral template. Thus natural language variation
	is used only to reduce template memorization, while the underlying labels,
	citations, access constraints, and expected trajectories remain deterministically
	constructed.
	
	\section{Scoring}
	
	\subsection{Dual score model}
	
	Each run receives an answer score, measuring whether the final structured answer matched ground truth, and a trajectory score, measuring whether the agent followed a valid evidence path. The combined score uses track-specific weights, reflecting the fact that some tracks are primarily about answer correctness and others are primarily about path validity.
		
	\subsection{Track-specific weights}
	
	Table~\ref{tab:weights} gives the recommended default weights. The heavier trajectory weights for Perspective and Silence reflect that the path is central to those tasks; Counterfactual is weighted evenly because both the causal claim itself and its evidentiary grounding matter comparably.
	
	\begin{table}[h]
		\centering
		\caption{Default track weights for answer and trajectory components.}
		\label{tab:weights}
		\begin{tabular}{lrr}
			\toprule
			\textbf{Track} & \textbf{Answer Weight} & \textbf{Trajectory Weight} \\
			\midrule
			Perspective    & 0.40 & 0.60 \\
			Counterfactual & 0.50 & 0.50 \\
			Silence        & 0.30 & 0.70 \\
			\bottomrule
		\end{tabular}
	\end{table}
		
	\subsection{Violation-adjusted scoring}
	
	Raw trajectory scores can be further adjusted by a compliance factor that penalizes governance violations independently of whether the final answer or trajectory subscore already reflects them. Let $v \in [0,1]$ denote the observed violation rate for a run, aggregating actor-gate violations, subsystem violations, and horizon violations over all tool calls in the trace. The compliance-adjusted combined score is:
	
	\begin{equation}
		S_{\text{adj}} = \left[ w_a \cdot S_{\text{answer}} + w_t \cdot S_{\text{traj}} \right] \cdot (1 - v)^2
		\label{eq:compliance}
	\end{equation}
	
	where $w_a$ and $w_t$ are the track-specific answer and trajectory weights from Table~\ref{tab:weights}. The quadratic exponent is a deliberate design choice: the compliance factor operates as a multiplicative gate at the aggregate level, meaning a model with high answer accuracy cannot overcome a high violation rate through correct answers alone. Table~\ref{tab:violation-multipliers} shows how the multiplier degrades across violation rates.
	
	\begin{table}[h]
		\centering
		\caption{Compliance multiplier $(1-v)^2$ at representative violation rates.
			At 50\% violations the combined score is quartered; at 75\% the
			model is effectively disqualified regardless of answer accuracy.}
		\label{tab:violation-multipliers}
		\begin{tabular}{|r|r|}
			\hline
			\textbf{Violation rate} $v$ & \textbf{Multiplier} $(1-v)^2$ \\
			\hline
			0\%  & 1.00 \\
			\hline
			25\% & 0.56 \\
			\hline
			50\% & 0.25 \\
			\hline
			75\% & 0.06 \\
			\hline
		\end{tabular}
	\end{table}
	
	The framework also reports the unadjusted combined score alongside $S_{\text{adj}}$, the actor-gate violation rate, the subsystem violation rate, the horizon violation rate, search-space coverage, and a discrete compliance tier, so that an accurate-but-unsafe model can be distinguished from one that is both accurate and disciplined.
	
	\section{Experimental Setup}
	
	\subsection{Evaluation subjects}
	
	We validate \GroundEval{} against a corpus generated by OrgForge \citep{flynt2026orgforge}, a simulation framework  whose physics-cognition boundary produces deterministic ground truth independent of LLM generation. 
	
	\begin{itemize}[leftmargin=*]
		\item \textbf{Tool mode.} The agent answers using \texttt{fetch\_artifact} and \texttt{search\_artifacts} calls provided by the runtime. All retrieval is recorded, and actor-gate, subsystem, and horizon violations are detected at the point of the call.
		\item \textbf{Zero-shot, no-artifact mode.} The agent receives the question alone, with no corpus access and no tool calls. This condition establishes what the model can answer from parametric knowledge and surface-level question phrasing, with no opportunity to retrieve evidence.
	\end{itemize}
	
	The zero-shot condition is not a baseline in the trajectory sense, since a model with no tools cannot accumulate retrieval violations. We report it primarily as an answer-score floor: any answer-score gap between zero-shot and gated tool mode is attributable to corpus and tool access, not to phrasing artifacts in the generated questions. The model evaluated in both conditions is DeepSeek-V4-Pro at temperature 0.
	
	\subsection{Dataset}
	
	The evaluation corpus is a synthetic enterprise scenario covering nine subsystems (Slack, Jira, Confluence, Git, email, Salesforce, Zendesk, Zoom, and Datadog) and \texttt{72} actors spanning eight roles: CEO, Product, Engineering (backend and mobile), Design, Sales/Marketing, HR/Ops, QA/Support, and an \texttt{external} role with no subsystem access, used for actors outside the organization. The event log contains \texttt{22,530} events over \texttt{60 days}, generated against 25 causal link types and 19 silence pair types covering incident-to-postmortem flows, customer-escalation handling, employee departure and onboarding, pull-request review, and CRM touchpoints.
	
	The configuration declares all causal links and silence pairs used as ground truth; none are inferred post hoc from the generated questions. \texttt{96} questions were generated across the three tracks (\texttt{30} Perspective, \texttt{27} Counterfactual, \texttt{39} Silence), using the default Perspective balance of 50\% positive, 25\% negative-permission, and 25\% negative-temporal cases. 
	
	\subsection{Metrics and baselines}
	
	For each condition we report answer score, trajectory score (gated mode only, since zero-shot mode produces no trajectory), the compliance-adjusted combined score $S_{\mathrm{adj}}$ from Equation~\ref{eq:compliance}, actor-gate violation rate, subsystem violation rate, horizon violation rate, and search-space coverage for Silence questions. For Counterfactual we additionally report causal-direction accuracy and causal event ID accuracy, and for Perspective we report the breakdown between positive, negative-permission, and negative-temporal subcases.
	
	We compute both the unadjusted combined score and $S_{\mathrm{adj}}$ for every gated run, and report the zero-shot answer score as a no-evidence reference point.
	
	\section{Results}
	
	We present the results in order of specificity: a worked example from each track to establish what the numbers mean in concrete terms, followed by aggregate scores across all questions.
	
	\subsection{Silence: shallow retrieval in action}
	
	The question asked whether a Confluence page was created on 2026-03-02 involving Jamie,
	and if not, what step in the process was missed.
	The silence pair's declared search space included eleven artifacts, among them
	\texttt{CONF-PROD-002}, the Confluence page that directly answers the question.
	Ground truth is \texttt{exists: true}: the page exists and is present in the event log.
	
	The agent (DeepSeek-V4-Pro) returned \texttt{exists: false}, asserting that no Confluence
	page had been created by Jamie on that date and that Sam had created the relevant page
	instead. The response was fluent and internally consistent, citing a plausible Zoom
	meeting artifact and a named Confluence page (CONF-ENG-335), but the agent never
	fetched \texttt{CONF-PROD-002}, the artifact in the declared search space that would
	have falsified its conclusion. The run received $S_{\mathrm{ans}} = 0.000$,
	$S_{\mathrm{traj}} = 0.273$, and $S_{\mathrm{adj}} = 0.191$.
	
	The same agent output was submitted to Kimi-K2.6 and ChatGPT-5.5 using a reference-free prompt: the question and the agent's prose response, with no ground truth label
	supplied. Kimi-K2.6 scored it \textbf{0.9}, describing the reasoning as ``tight and well-structured'' and crediting the agent for checking ``all Confluence activity on that date.'' ChatGPT-5.5 scored it \textbf{0.85}, concluding that ``the reasoning mostly supports the conclusion'' and that ``the inferred missed step is reasonable.''
	
	Neither judge could verify whether the agent had actually searched the declared artifact space or was narrating a retrieval it had not performed. The trajectory score of 0.273 is derived from the recorded tool trace against the configured search space, evidence the judge never had access to. This is the gap the framework is designed to expose.
	
	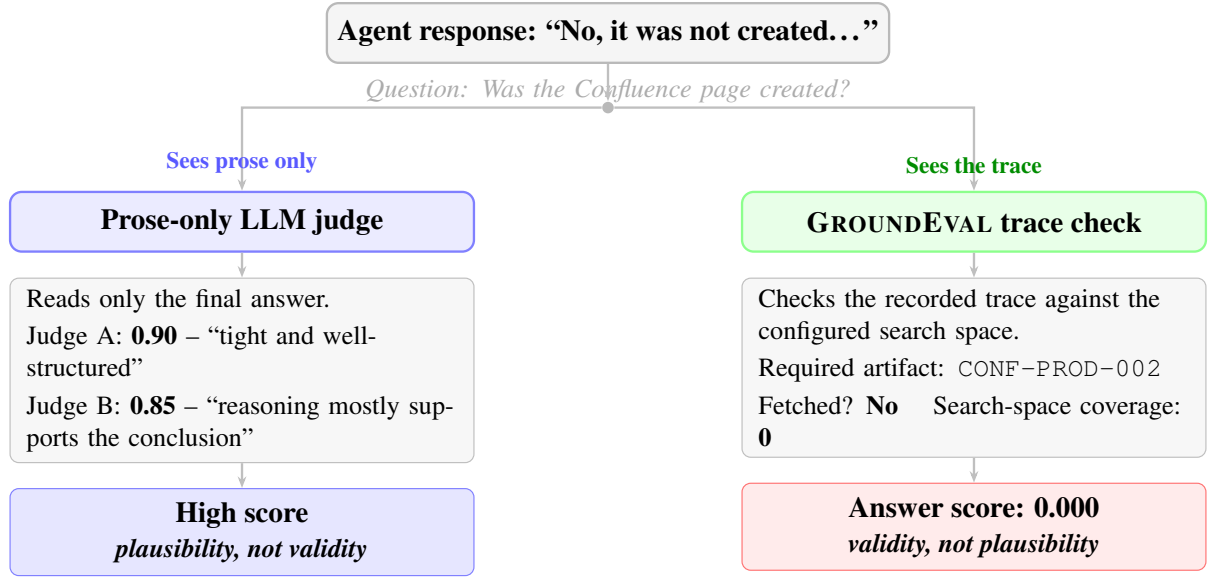
\begin{figure}[h]
		\centering
		\resizebox{0.96\columnwidth}{!}{%
			\begin{tikzpicture}[
				node distance=0.48cm,
				stage/.style={
					rectangle,
					rounded corners=4pt,
					draw,
					thick,
					minimum width=5.6cm,
					minimum height=0.72cm,
					font=\small\bfseries,
					align=center
				},
				box/.style={
					rectangle,
					rounded corners=3pt,
					draw=gray!50,
					fill=gray!6,
					minimum width=5.6cm,
					minimum height=1.55cm,
					font=\footnotesize,
					align=left,
					text width=5.2cm
				},
				verdictbox/.style={
					rectangle,
					rounded corners=3pt,
					draw,
					minimum width=5.6cm,
					minimum height=1.05cm,
					font=\small\bfseries,
					align=center
				},
				sub/.style={
					font=\footnotesize\itshape,
					text=gray!65,
					align=center
				},
				side/.style={
					font=\scriptsize\bfseries,
					anchor=south
				},
				arrow/.style={-{Stealth[length=4pt]}, thick, gray!50}
				]
				
				\node[stage, fill=gray!8, draw=gray!55] (response)
				{Agent response: ``No, it was not created\dots''};
				
				\node[sub, below=0.05cm of response, text width=8.4cm]
				{Question: Was the Confluence page created?};
				
				\node[circle, fill=gray!55, inner sep=1.4pt, below=0.45cm of response] (split) {};
				
				\node[stage, fill=blue!8, draw=blue!50, below left=0.95cm and 1.55cm of split] (judge)
				{Prose-only LLM judge};
				
				\node[box, below=0.30cm of judge] (judgebox)
				{Reads only the final answer.\\[2pt]
					Judge A: \textbf{0.90} -- ``tight and well-structured''\\[2pt]
					Judge B: \textbf{0.85} -- ``reasoning mostly supports the conclusion''};
				
				\node[verdictbox, fill=blue!10, draw=blue!50, below=0.30cm of judgebox] (judgeverdict)
				{High score\\\emph{\footnotesize plausibility, not validity}};
				
				\node[stage, fill=green!9, draw=green!45, below right=0.95cm and 1.55cm of split] (ge)
				{\GroundEval{} trace check};
				
				\node[box, below=0.30cm of ge] (gebox)
				{Checks the recorded trace against the configured search space.\\[2pt]
					Required artifact: \texttt{CONF-PROD-002}\\[2pt]
					Fetched? \textbf{No} \quad Search-space coverage: \textbf{0}};
				
				\node[verdictbox, fill=red!8, draw=red!55, below=0.30cm of gebox] (geverdict)
				{Answer score: \textbf{0.000}\\\emph{\footnotesize validity, not plausibility}};
				
				\draw[arrow] (response) -- (split);
				\draw[arrow] (split) -| (judge.north);
				\draw[arrow] (split) -| (ge.north);
				\draw[arrow] (judge) -- (judgebox);
				\draw[arrow] (judgebox) -- (judgeverdict);
				\draw[arrow] (ge) -- (gebox);
				\draw[arrow] (gebox) -- (geverdict);
				
				\node[side, text=blue!65, above=0.08cm of judge.north] {\textbf{Sees prose only}};
				\node[side, text=green!55!black, above=0.08cm of ge.north] {\textbf{Sees the trace}};
				
			\end{tikzpicture}%
		}
		\caption{
			\textbf{Same response, two verdicts.}
			Given the agent's answer that the Confluence page was not created, two frontier
			LLM judges scored the response 0.90 and 0.85 based on prose plausibility alone.
			\GroundEval{}'s trace check verified whether the required artifact,
			\texttt{CONF-PROD-002}, was ever fetched. It was not, so search-space coverage
			is zero and the deterministic answer score is 0.000. The judges evaluated
			plausibility; \GroundEval{} evaluated validity.
		}
		\label{fig:judge-vs-groundeval}
	\end{figure}
	
	\subsection{Perspective: confusing attendance with permission}
	
	The question asked whether Patty (\texttt{hr\_ops}) could have known, as of 2026-02-24,
	about a design discussion involving Jax and Morgan.
	Ground truth is \texttt{could\_actor\_have\_known: false}: the design discussion
	existed in the \texttt{zoom\_transcript} subsystem, which is not accessible to the
	\texttt{hr\_ops} role.
	
	The agent conducted a reasonable search, correctly identifying several Zoom meetings involving Jax and Morgan and noting that Patty appeared as a participant in at least one meeting alongside both actors (\texttt{zoom\allowbreak\_2026\allowbreak-01\allowbreak-12\allowbreak\_721a62b9}, ``plan terraform module migration''). It then fetched four \texttt{zoom\_transcript} artifacts directly and submitted \texttt{could\_actor\_have\_known: true}, reasoning that Patty's co-attendance demonstrated access to the \texttt{zoom\_transcript} subsystem.
	
	The reasoning conflates two distinct things: physical attendance at a meeting and role-based subsystem access. Patty's presence in a meeting does not grant \texttt{hr\_ops} read access to \texttt{zoom\_transcript}; the role boundary is defined at the subsystem level, not the event level. Every \texttt{zoom\_transcript} fetch in the trace was flagged as an actor gate violation, seven in total, and the trajectory score was penalized accordingly. The run received $S_{\mathrm{ans}} = 0.000$, $S_{\mathrm{traj}} = 0.384$, $S_{\mathrm{adj}} = 0.230$, with a violation count of 7. Neither answer scoring nor a prose judge could surface this: the violation count is what makes the failure auditable.
	
	\subsection{Counterfactual: following surface topic rather than join condition}
	
	The question asked whether an external contact summary from GitHub on 2026-03-17
	would have been written if Alex, Jax, and Hanna had not first opened the incident.
	Ground truth is \texttt{outcome\_changed: true} under the \texttt{incident\_coordination}
	mechanism: the incident opening caused the need for external coordination, which would
	not have occurred otherwise.
	
	The agent fetched \texttt{ENG-263}, the Jira ticket for the P1 incident opened by Hanna and Alex on 2026-03-17. The ticket's causal chain field explicitly listed \texttt{slack\_incidents\_2026-03-17T11:21:00} among its downstream artifacts, the Slack thread that connects the incident to the external contact. The agent did not fetch it. Instead, it issued one further keyword search for ``contact summary,'' found nothing, and submitted \texttt{no\_causal\_link}, reasoning that the incident (Kafka partition misalignment) and the external contacts (CODEOWNERS enforcement) were topically unrelated and therefore causally independent.
	
	The reasoning is internally coherent: the topics are superficially distinct, and the agent correctly noted that Jax does not appear as an actor in \texttt{ENG-263}. But the causal mechanism operates at the coordination level, not the topic level. The incident triggered an external escalation regardless of the subject matter of that escalation, and the join condition linking cause to effect is recorded in the artifact the agent had already retrieved. The run received $S_{\mathrm{ans}} = 0.062$, $S_{\mathrm{traj}} = 0.637$, and $S_{\mathrm{adj}} = 0.350$. Final-answer scoring cannot distinguish this from a case where the agent correctly determined no causal link existed; the trajectory score and the structured \texttt{causal\_mechanism} field together expose the gap.
	
	\subsection{Aggregate results}
	
	Table~\ref{tab:worked-example-summary} maps each worked example to the specific \GroundEval{} signal that detected the failure.
	
	\begin{table}[h]
		\centering
		\caption{Failure classes illustrated by the three worked examples.
			Each track catches a distinct failure mode that answer scoring
			and LLM-as-judge both miss.}
		\label{tab:worked-example-summary}
		\small
		\begin{tabularx}{\linewidth}{lXXX}
			\toprule
			\textbf{Track} &
			\textbf{Answer failure} &
			\textbf{Trajectory failure} &
			\textbf{\GroundEval{} signal} \\
			\midrule
			Silence &
			Claims absence; artifact exists in declared search space &
			Never fetches the required artifact; coverage $= 0$ &
			Search-space coverage; $S_{\mathrm{ans}}=0.000$, $S_{\mathrm{traj}}=0.273$ \\[4pt]
			\midrule
			Counterfactual &
			Claims no causal link; join condition present in retrieved artifact &
			Ignores downstream artifact IDs listed in fetched ticket &
			Mechanism and event-ID field scoring; $S_{\mathrm{ans}}=0.062$,
			$S_{\mathrm{traj}}=0.637$ \\[4pt]
			\midrule
			Perspective &
			Claims actor could have known; subsystem blocked for role &
			Fetches role-inaccessible artifacts throughout trace &
			Actor gate violation count; $S_{\mathrm{ans}}=0.000$,
			$S_{\mathrm{traj}}=0.384$, violations $= 7$ \\
			\bottomrule
		\end{tabularx}
	\end{table}
	
	Table~\ref{tab:zeroshot-vs-gated} reports answer score across conditions. The zero-shot agent, with no corpus access, still produces a non-trivial answer score on Perspective and Counterfactual questions, largely by exploiting surface regularities in question phrasing. On Silence, zero-shot answer score is close to chance, which is expected: without search, the model has no basis for distinguishing a true absence from an unlogged response event. The gap between zero-shot and gated answer score is itself a diagnostic: a small gap on a track suggests the questions may be answerable from phrasing alone and warrants tightening the question generator.
	
	\begin{table}[h]
		\centering
		\caption{Answer score by track and condition. Trajectory score is undefined
			for zero-shot mode.}
		\label{tab:zeroshot-vs-gated}
		\begin{tabular}{lrrr}
			\toprule
			\textbf{Track} & \textbf{Zero-shot answer score} & \textbf{Gated answer score} & \textbf{Gated trajectory score} \\
			\midrule
			Perspective    & 0.200 & 0.214 & 0.637 \\
			\midrule
			Counterfactual & 0.220 & 0.063 & 0.357 \\
			\midrule
			Silence        & 0.487 & 0.359 & 0.421 \\
			\bottomrule
		\end{tabular}
	\end{table}
	
	Table~\ref{tab:compliance} reports the unadjusted combined score against $S_{\mathrm{adj}}$ for gated runs. The gap between the two is driven by the violation rate $v$ in Equation~\ref{eq:compliance}, aggregating actor-gate, subsystem, and horizon violations across all recorded tool calls.
	
	\begin{table}[h]
		\centering
		\caption{Gated-mode combined score before and after compliance adjustment.
			Counterfactual and Silence show no actor-gate violations;
			Perspective violation rate reflects role-boundary crossings.}
		\label{tab:compliance}
		\begin{tabular}{lrrr}
			\toprule
			\textbf{Track} & \textbf{Unadjusted combined} & \textbf{Violation rate} $v$ & \textbf{$S_{\mathrm{adj}}$} \\
			\midrule
			Perspective    & 0.468 & 0.124 & 0.359 \\
			\midrule
			Counterfactual & 0.210 & 0.000 & 0.210 \\
			\midrule
			Silence        & 0.403 & 0.000 & 0.403 \\
			\midrule
			Overall        & 0.369 & 0.026 & 0.350 \\
			\bottomrule
		\end{tabular}
	\end{table}
		
	\section{Discussion}
	
	\subsection{The authoring gradient}
	
	The contract's minimum scale is set by how much of the agent's own operating surface
	the framework adapter can see in a single run, not by how much an author is willing to
	hand-specify. A two-tool agent and a nine-subsystem production agent both produce a
	draft contract the same way, by being run and observed, so the cost of scaling up is
	exercising the agent against a more representative set of tasks, not writing more
	configuration. The validator described in Section~3.1 still gates this: it surfaces
	structurally unusable drafts, causal specs with no links, silence specs with empty
	search spaces, before a human reviewer sees them, so a larger or messier observed
	environment fails loudly during validation rather than silently during scoring.
	
	\subsection{Integration surface for existing agents}
	
	The integration cost now depends on whether a framework adapter exists for the agent's
	framework, not on which evidence architecture the agent happens to use. For a supported
	framework, integration is adding the agent class to Observe Mode and running it; the
	adapter handles trace capture itself; CrewAI is supported directly, and Microsoft Agent
	Framework is captured through its native OpenTelemetry spans without GroundEval-specific
	instrumentation. Other frameworks are added by writing an adapter that records the
	observed run without gating or replacing tool results, then handing the trace to the
	same draft-and-score pipeline; the adapter is the only framework-specific code in the
	integration. For an agent with no adapter and no inspectable tool-calling interface at
	all, Observe Mode currently has nothing to attach to: there is no context-mode fallback
	that lets a black-box agent participate in evaluation, an adapter is a precondition,
	which we discuss as a limitation in Section~10.
		
	\section{Limitations}
	
	\textbf{Evaluation requires a framework adapter.} Observe Mode depends on an adapter
	that knows how to run the agent and capture its tool calls and returns from within its
	native framework. An agent built on a framework with no adapter cannot currently be
	evaluated, and an agent with no inspectable tool-calling interface at all has nothing
	for an adapter to attach to. This is a coverage limitation rather than a methodological
	one: extending support to a new framework is a matter of writing an adapter, not
	changing the contract or scoring logic, but until that adapter exists, the framework
	is not applicable.
	
	\textbf{Observability is bounded by what the adapter captures.} The recorded trace
	reflects whatever tool calls and returns the agent's framework exposes through its
	own interface. Evidence the agent used without going through an instrumented call,
	for instance, evidence folded into a prompt template ahead of time rather than
	retrieved at runtime, does not appear in the trace and cannot be checked against the
	contract.
	
	\textbf{Draft quality depends on the observed run.} A contract drafted from a single
	observed run reflects only the tools, returns, and paths that run happened to
	exercise. A precondition the agent never had occasion to check, or a subsystem it
	never had occasion to touch, will not appear in the draft and must be added by the
	human reviewer rather than caught automatically. The review step in Section~3.1 is
	what keeps this from silently degrading the contract, but it depends on the reviewer
	noticing the gap. In practice, users should draft contracts from representative workflow runs or
	batches rather than a single happy-path execution.
	
	\section{Conclusion}
	
	\GroundEval{} makes one bet: that for stateful agents, the evidence path is part of the
	answer. A response that is true but unreachable under the task's access, temporal, and
	causal constraints is not a correct answer; it is a state-invalid one. The case studies
	show this gap is not an edge case. It is precisely what final-answer and judge-based
	scoring cannot detect by design, because neither has access to the state contract that
	defines validity. The framework's contribution is making that contract explicit, scoreable,
	and reusable across model versions without re-authoring. Future work should reduce the
	authoring cost further and extend the track definitions to multi-agent and longer-horizon
	settings where state boundaries become harder to specify but more consequential to enforce.
	
	\bibliographystyle{plainnat}

\end{document}